# Show, Attend and Interact: Perceivable Human-Robot Social Interaction through Neural Attention Q-Network

Ahmed Hussain Qureshi, Yutaka Nakamura, Yuichiro Yoshikawa and Hiroshi Ishiguro

*Abstract*— For a safe, natural and effective human-robot social interaction, it is essential to develop a system that allows a robot to demonstrate the perceivable responsive behaviors to complex human behaviors. We introduce the Multimodal Deep Attention Recurrent Q-Network using which the robot exhibits human-like social interaction skills after 14 days of interacting with people in an uncontrolled real world. Each and every day during the 14 days, the system gathered robot interaction experiences with people through a hit-and-trial method and then trained the MDARQN on these experiences using end-to-end reinforcement learning approach. The results of interaction based learning indicate that the robot has learned to respond to complex human behaviors in a perceivable and socially acceptable manner.

## I. INTRODUCTION

Human-robot social interaction (HRSI) is an emerging field with an aim of bringing robots into our social world as our companions. For robots to coexist with humans, it is crucial for them to predict human intentions in order to respond to each and every one of the countless and complex human behaviors with utmost propriety [1].

Human intention prediction is a challenging task [2] as it depends on many intention depicting factors such as human walking trajectory, face expression, gaze direction, body movement or any ongoing activity. Therefore, programming a robot which can interpret and respond to complex human behaviors based on their intentions is notoriously hard. To solve this challenge we believe that it is essential to augment robots with a self-learning architecture [3] which enables them to learn social interaction skills from high-dimensional interaction experiences automatically.

Recent advancements in machine learning has combined deep learning with reinforcement learning and has led to the development of Deep Q-Network (DQN) [4]. DQN utilizes deep convolutional neural network [5] for the approximation of Q-learning's action-value function. DQN has demonstrated its ability to play arcade video games at human and superhuman level by learning, through hit and trial method, from high dimensional visual data. However, the applicability of DQN to real world human-robot interaction problem was not explored until we, recently, proposed the multimodal deep Q network (MDQN) [6] for HRSI.

A. H. Qureshi, Y. Nakamura, Y. Yoshikawa and H. Ishiguro are with Department of System Innovation, Graduate School of Engineering Science, Osaka University, 1-3 Machikaneyama, Toyonaka, Osaka, Japan. {qureshi.ahmed, nakamura, yoshikawa, ishiguro}@irl.sys.es.osaka-u.ac.jp
A. H. Qureshi, Y. Yoshikawa and H. Ishiguro are also with JST ERATO ISHIGURO Symbiotic Human-Robot Interaction Project.

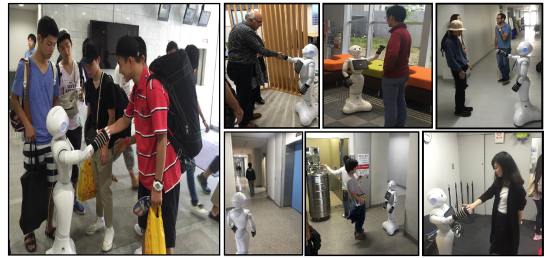

Fig. 1: Robot learning social interaction skills from people.

MDQN uses dual stream convolutional neural networks for action-value function approximation. The dual stream structure processes the grayscale and depth images independently, and the Q-values from both streams are fused together for choosing the best possible action in the given scenario. By using MDQN the robot learned to greet people after 14 days of hit and trial method based interaction with people at different public places such as a cafeteria, common rooms, department entrance, etc (as shown in figure 1). The robot could perform only one of the four actions for an interaction and the action were waiting, looking towards human, waving hand and handshaking. Results showed that the robot augmented with MDQN learned to choose appropriate actions in the diverse real world scenarios. However, in [6], the robot actions lacked perceivability as the robot could not indicate its attention. The research in [7] highlights that humans show more willingness to interact with a robot that can indicate its attention than with a robot that cannot. Therefore, in this paper we propose a Multimodal Deep Attention Recurrent Q-Network (MDARQN) which adds perceivability to robot actions through a recurrent attention model (RAM) [8].

RAM enables the Q-network to focus on certain parts of the input image instead of processing it entirely at a fine scale. This region selection reduces the number of training parameters as well as the computational operations. Beside computational benefits, RAM provides information about where MDARQN is looking at, while taking any decision. In the proposed work, we utilize this visual attention information by RAM for realizing perceivable HRSI.

## II. RELATED WORK

The challenge of modeling responsive robot behaviors for a wide diversity of complex human behaviors has gained interest of many researchers. Recently, work by Lee et al. [9], Amor et al. [10] [11] and Wang et al. [2] addresses the said challenge. The proposed work in [9] [10] [11] uses a motion capture system for recording the interaction between two persons and the responsive robot behavior is learned from the recorded data by imitating the behavior of human interaction partners. We believe that the motion capture system does not yield natural interaction behaviors as the participants are required to wear special skin-tight dress together with the track-able makers. In [2], the authors proposed a probabilistic graphical model using which the human intentions are inferred from the observed body movements. However, as mentioned earlier, intention prediction relies on various intention depicting factors, thus, inferring intention from body movements alone is not sufficient. Furthermore, aforestated prior art considers only one human interaction partner for the robot at any time but in proposed research the robot operates in natural uncontrolled environment where it can be approached by any number of people. The quest of an efficient intention predictor has also led to the deep learning based method [12]. In [12], the authors used video data for training an intention predictor. However, in our work, the interactive behavior perception is crucial because the robot is an active agent in the environment and hence, it can alter the human intention by taking any action. Therefore, the robot needs to interpret the human behavior and its own existence under the human social norms before making any decision. Recently, we proposed MDQN [6] for interactive behavior perception. The robot augmented with MDQN does not perform perceivable interaction with people because of no attention mechanism. In our proposed work we utilize recurrent attention models (RAM) for perceivable robot actions. So far, recurrent attention models (RAM) have been applied successfully to various tasks such as object tracking [13], image classification [13], machine translation [14], and image captioning [15] . The research in [8], integrate RAM into DQN and surpasses the previous performance of DQN on some of the Atari games. RAM provides insight into the behavior of the Q-network and in our proposed work, we utilize this insight for driving the robot attention onto the regions of an input scene where the Q-network is focusing on while making any decision. To the best of our knowledge, the applicability of RAM for the perceivable HRSI has not been explored yet.

## III. BACKGROUND

In this work, the human-robot interaction problem is formalized as standard reinforcement Q-learning task in which the agent interacts with an environment $E$ through an action $a \in \mathcal{A}$ and gets a scalar reward $r$, where $\mathcal{A} = \{1, \cdots, K\}$ is the set of all legal actions.

The Q-learning agent learns an *action-value* function which maps an input state $s$ to an action $a$ under a policy $\pi$ i.e., $Q^\pi(s,a) = \mathbb{E}[R_t|s_t = s, a_t = a, \pi]$. The objective of a Q-agent is to maximize the expected total return $R_t = \sum_{t'=t}^{T} \gamma^{t'-t} r^{t'}$, where $\gamma : [0,1]$ is a discount factor, $r$ is an immediate reward and $T$ is the terminal step. The maximum achievable expected total return under policy $\pi$ is determined by an optimal *action-value* function $Q^*(s,a) = \max Q^\pi(s,a)$, and this function obeys a fundamental Bellman relation $Q^*(s,a) = \mathbb{E}[r + \gamma \max_{a_{t+1}} Q^*(s_{t+1}, a_{t+1})|s_t, a_t]$. The Bellman relation can be interpreted as: for a sequence $s_{t+1}$ at next time-step if the action-value function $Q^*(s_{t+1}, a_{t+1})$ is deterministic for all possible actions $\mathcal{A}$ then the optimal policy is to choose an action $a_{t+1}$ which maximizes the expected value of $r + \gamma Q^*(s_{t+1}, a_{t+1})$.

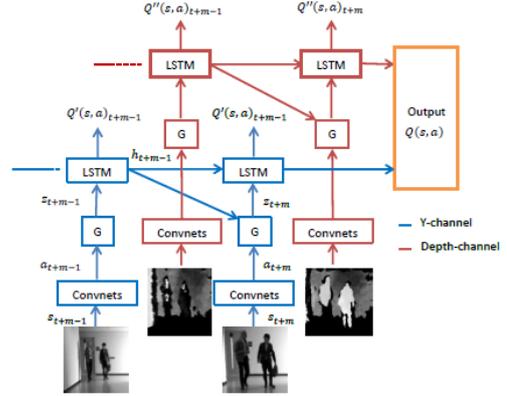

Fig. 2: Multimodal Deep Attention Recurrent Q-Network

In practical Q-learning, the action-value function is approximated by a function estimator such as neural networks i.e., $Q(s,a) \approx Q(s,a,\theta)$ and the parameters of an estimator are adjusted iteratively towards the Bellman target. Recently, a Deep Q-Network (DQN) is introduced which uses a deep convolutional neural networks (convnets) as a function approximator and the parameters of convnets are trained by minimizing the following loss function:

$$L_t(\theta) = \mathbb{E}\left[\left(r + \gamma \max_{a_{t+1}} Q(s_{t+1}, a_{t+1}; \theta^-) - Q(s,a;\theta)\right)^2\right] \quad (1)$$

The DQN network uses two Q-networks for minimizing the loss function (Eq 1) i.e., the Bellman target $B_t = r + \gamma Q^*(s_{t+1}, a_{t+1}; \theta^-)$ is computed by a target Q-network with old parameters $\theta^-$ while the training Q-network maintains the recently updated parameters $\theta$. The old parameters $\theta^-$ are updated to current parameters after every C steps. The gradient of loss function (Eq 1) with respect to parameters $\theta$ takes the following form:

$$\nabla L_t(\theta) = \mathbb{E}\left[(B_t - Q(s,a;\theta)) \nabla_\theta Q(s,a;\theta)\right] \quad (2)$$

In addition to maintaining two Q-networks, DQN also uses experience replay [16] for training Q-networks. Finally, DQN follows an $\epsilon$-greedy strategy for interacting with Atari emulator i.e., with probability $1 - \epsilon$ the agent takes greedy action by exploiting the Q-network while with probability $\epsilon$ the agent randomly picked an action $a \in \mathcal{A}$ for exploration.

## IV. THE PROPOSED MDARQN

In this section, we describe our proposed neural model i.e., MDARQN using which the robot learns to do perceivable HRSI. The MDARQN architecture comprises of two streams of identically structured neural Q-networks, one for processing the grayscale frames while other for processing the depth frames. Each of these neural Q-network streams is trained independently of each other. Since two Q-network streams are identical, and trained independently, therefore, for simplicity; we only discuss the structure of a single stream of the dual stream Q-network. Each stream consists of three neural models: 1) Convnets; 2) Long Short-term Memory (LSTM) network; and 3) Attention network (G). The rest of the section explains these three neural models and the flow of information between them (as also shown in figure 2).

*1) Convnets:* The convnets take pre-processed visual frame as an input at each time-step and transform it into $L$ feature vectors, each of which provides D-dimensional representation of a part of an input image i.e, $a_t = \{\mathbf{a}_t^1, \cdots, \mathbf{a}_t^L\}, \mathbf{a}_t^l \in \mathbb{R}^D$. This feature vector is taken as an input by the attention network for generating the annotation vector $z \in \mathbb{R}^D$.

*2) LSTM:* We employ the following implementation of LSTM network:

$$\begin{pmatrix} i_t \\ f_t \\ o_t \\ g_t \end{pmatrix} = \begin{pmatrix} \sigma \\ \sigma \\ \sigma \\ \tanh \end{pmatrix} M \begin{pmatrix} h_{t-1} \\ z_t \end{pmatrix} \quad (3)$$

$$c_t = f_t \odot c_{t-1} + i_t \odot g_t \quad (4)$$

$$h_t = o_t \odot \tanh(c_t) \quad (5)$$

where $i_t$, $f_t$, $o_t$, $c_t$, and $h_t$ correspond to the input, forget, output, memory and hidden state of the LSTM, respectively. Let $d$ be the dimensionality of all LSTM states and matrix $M : \mathbb{R}^a \to \mathbb{R}^b$, in equation 3, is an affine transformation of trainable parameters with dimension $a = d + D$ and $b = 4d$. As shown in equation 3-5, the LSTM network takes the annotation vector $z_t \in \mathbb{R}^D$, previous hidden state $h_{t-1}$, and the previous memory state $c_{t-1}$ as an input in order to produce the next hidden state $h_t$. This hidden state $h_t$ is given to the attention network $G$ and to the linear output layer for generating the annotation vector $z_{t+1}$ at next time step and for providing the output Q-value for each of the legal actions, respectively.

*3) Attention network:* The attention network generates the dynamic representation, called annotation vector $z_t$, of the corresponding parts of an input image at time $t$. The attention mechanism $\phi$, a multilayer perceptron, takes a D-dimensional L feature vectors $a_t$ and a previous hidden state $h_{t-1}$ of the LSTM network as an input for computing the positive weights $\beta_t^l$ for each location $l$. The weights $\beta_t^l$ are computed as follow:

$$\beta_t^l = \frac{\exp(\alpha_t^l)}{\sum_{k=1}^{L} \exp(\alpha_t^k)}; \text{ where } \alpha_t^l = \phi(a_t^l, h_{t-1}) \quad (6)$$

The annotation vector $z_t$ is computed as $z_t = \sum_{l=1}^{L} \beta_t^l a_t^l$. This annotation vector is used by LSTM for computing next hidden state.

There are two type of attention network [15] in the literature: the soft and hard attention network. The attention network used in MDARQN is the soft attention network and unlike hard attention network, it fully differentiable and deterministic.

Since each of the streams of the MDARQN model is fully differentiable, therefore, each network stream is trained by minimizing the general loss function (Equation 1) through a standard back-propagation method. Finally, output from the two streams are fused together for taking a greedy action as shown in the figure 2. For the fusion, the output Q-values from each Q-network stream are first normalized and then these normalized Q-values from each stream are averaged together to generate output Q-values of MDARQN. The greedy action is then taken by picking the action which has a highest Q-value from these fused Q-values.

V. IMPLEMENTATION DETAILS

This section outlines the implementation details of the proposed project. The MDARQN code was built on the baseline [4] [8] and is implemented in torch/lua[1]. The robot side programming is done in python. The system used for training MDARQN has 3.40GHz×8 Intel Core i7 processor with 32 GB RAM and GeForce GTX 980 GPU. The rest of the section explains various modules of the project.

*A. Robotic system*

A Pepper robot[2] was used for the proposed project. Out of many built-in sensors of the Pepper, we only use a 2-D camera located on robot's forehead and an ASUS Xtion 3-D sensor located behind robot eyes for the grayscale and depth images, respectively. The 2-D camera and the 3-D sensor were operated at 10 fps with 320×240 resolution. In addition to visual sensors, we also equip Pepper's right hand with FSR touch sensor which detects if the handshake has happened or not and this handshake detection forms the basis for our reward function (as discussed later). For aesthetic reasons we also hide the touch sensor under the woolen gloves as can be seen in figure 1.

*B. Robot actions with attention*

In order to ensure perceivable HRSI, we utilize the annotation vector $z$ given by attention network $G$ for attention steering of the robot. This attention steering is done as follow.

*Attention steering for greedy actions:* The images used by the MDARQN has dimensions $198 \times 198$. We divide horizontal and vertical axis of the input image into five sub-regions based on the author's defined thresholds. The horizontal axis is divided into the left most, left, center, right, right most regions while the vertical axis is divided into top most, top, center, bottom and bottom most regions. The indicators $I^x \in \{-2, -1, 0, 1, 2\}$ and $I^y \in \{-2, -1, 0, 1, 2\}$ indicate these sub-regions of horizontal and vertical axis, respectively, starting from -2 which corresponds to the left/top most region. The robot attention mechanism uses annotation vector $z_t$ to extract the pixel location on the input image, we call it attention mark, where the MDARQN pays the maximum attention. The attention mark and author's defined thresholds are then used to determine the indicator values $I = \{I^x, I^y\}$. The indicator $I$ and robot's actual position indicator $I^a = \{I^a_x, I^a_y\}$ is then used to compute the next attention location. The value of robot's actual position indicator at the given time-step is computed relative to the actual location at previous time-step and it is calculated as follow:

$$I^a_t = \begin{cases} 0, \text{if } I + I^a_{t-1} > 2 \text{ or } I + I^a_{t-1} < -2 \\ I + I^a_{t-1}, \text{otherwise} \end{cases} \quad (7)$$

---
[1] http://torch.ch/
[2] //www.aldebaran.com/en/cool-robots/pepper/find-out-more-about-pepper

The robot actual position is initialized to its central location i.e., $I^a_0 = 0$. The motion of the robot in a real world is determined by $\omega = \{\omega_x, \omega_y\}$ where $\omega_x$ controls the rotation of the robot body while $\omega_y$ controls the robot's head projection. The $\omega$ is computed as follow:

$$\{\omega_x, \omega_y\} = \{s\theta_1, s\theta_2\} \text{ where } s = I^a_t - I^a_{t-1} \quad (8)$$

The value of $\theta_1$ and $\theta_2$ are $\pi/6$ and $\pi/9$, respectively. It should be noted that it is important to bind the robot motion to predefined regions. As in the proposed work we utilized embedded visual sensors in the robot which have limited field of view and are mobile due to the robot motion. Hence, restricted motion allows a safe localization of robot in public environments.

*Attention steering for non-greedy actions:* Since during non-greedy robot's behavior, the system randomly picks an action from the set of legal actions, therefore, it is necessary to equip the robot with another attention system that facilitates it's interaction with humans. This function instills the awareness into the robot and makes it sensitive to the stimulus coming from the real world. The stimuli used are the sound and the movement detection. In case robot senses any stimulus, it looks for human at the stimulus origin. If there is not any human, the robot returns to its previous orientation but otherwise it tracks the human with its head in order to engage them for an interaction.

After attending, the robot executes a chosen action. The rest of this section describes the implementation details of these four legal actions, i.e., waiting, looking towards humans, waving its hand and hand shaking with a human.

*Wait:* For this action, during a greedy policy, the robot does nothing other than attending to the attention location. However, in case of non-greedy policy, the robot randomly moves its head within allowable range of head pitch and head yaw.

*Look towards human:* During this action, if there is human, the robot tracks the person with its head. If this action is being performed under a greedy policy then the robot tracks the human within a narrow field in order to avoid any desynchronization with the greedy attention mechanism.

*Wave hand:* During this, the robot waves its hand and says *Hello*.

*Handshake:* For performing a handshake, the robot lifts its right hand up to a certain height and then waits for a few seconds. If FSR touch sensor detects the touch then the robot grabs the person's hand otherwise robot brings its hand down to the default position.

### C. Reward function

Handshake detection through touch sensor forms the baseline of our reward function. The robot gets the reward of 1 and -0.1 on the successful and unsuccessful handshake, respectively. Furthermore, the reward of value 0 is given on actions other than handshake. The handshake is successful if the human and robot actually shake each others hand while it is unsuccessful when robot attempts to do a handshake but the handshake does not happen.

### D. Model Architecture

This section provides the architecture details of MDARQN model. The MDARQN consist of two streams: the Y-channel and the Depth-channel stream for processing the grayscale and depth images, respectively. Since, the structure of both streams are identical, therefore, we only discuss one of the streams.

The convolutional neural network consists of four convolution layers each of which is followed by a non-linear rectifier function. The input dimension to the CNN is $1 \times 198 \times 198$. The convolution layer 1, 2, 3 and 4 convolves 16 filters of $9 \times 9$, 32 filters of $8 \times 8$, 64 filters of $7 \times 7$ and 256 filter of $6 \times 6$, respectively. The stride of convolution 1, 2, 3 and 4 are 3, 2, 2 and 1 respectively. The CNN outputs 256 feature maps of dimension $7 \times 7$. The output feature maps from CNN are given to the attention network which takes 49 vectors each of size 256. To be consistent with attention network, the LSTM network also has 256 units. To generate Q-values for the four set of actions, the output of the LSTM is transformed to four units through a linear layer preceded by non-linear rectifier unit.

### E. Training dataset, data augmentation and pre-processing

We double the training dataset through two data augmentation techniques: 1) Random cropping of the input image of size $320 \times 240$ to the size suitable for the model i.e., $198 \times 198$; 2) Mirroring the input image and then cropping it randomly to the size $198 \times 198$. The total training data collected during 14 days of experiment comprise of 111,504 grayscale and depth frames. After data augmentation, the number of grayscale and depth frames grows to 223,008. To prepare an input for the MDARQN, the eight most recent depth and grayscale frames, of each time step, are stacked together to form an input for the Y-channel and the Depth-channel of the MDARQN, respectively.

### F. Training procedure

We present a training procedure which comprise of two phases, the *data generation phase* and *learning phase*.

*1) Data generation phase:* In this phase the agent interacts with an environment for generating interaction experiences $e$. At time $t$, the environment provides an observation state $s_t$, the agent after observing a state $s_t$ takes an action $a_t$ using $\epsilon$-greedy policy, the environment in return provides the scalar reward $r_t$ and the next state $s_{t+1}$. The interaction experience $e_t = \{s_t, a_t, r_t, s_{t+1}\}$ is then stored into a replay buffer $\mathcal{M}$ for experience replay during the learning phase.

This cycle of generating data keeps on repeating until the terminal state $T$ is achieved. The replay buffer stores $N$ most recent interaction experiences.

*2) Learning phase:* During this phase, the agent feeds on the replay memory $\mathcal{M}$ for training the MDARQN $Q(s, a; \theta)$ by minimizing loss function (Equation 1). Like DQN training, we also maintain two MDARQN i.e., the target network and current network with old parameters $\theta^-$ and new $\theta$ parameters, respectively.

In the propose work, the MDARQN agent was trained for 14 days. Every day, the robot interacted with people for some time period $T$ in order to generate a data (*data-generation phase*). After $T$ time period, the robot went to rest position and the learning phase began. It should be noted that the proposed training method is different from the DQN training procedure [4]. In DQN training, after filling a memory buffer with $n$ experiences, the Q-network is trained on a minibatch after collecting each and every interaction experience $e$. This training of Q-network after every single interaction experience adds a delay between the agent's interaction with an environment at time $t$ and at time $t + 1$. In [4], the environment for the DQN is an Atari emulator which is somehow controllable. From the word controllable we mean that during the DQN training, the Atari environment halts and it waits for the DQN-agent to execute its next action. In our proposed work, the environment is real, uncontrollable and it requires the MDARQN agent to interact with people. Therefore any significant delay while robot is in the field for interaction with the people is unacceptable. Hence, we divided the training procedure into two phases.

### G. Experiment details and hyper-parameters

We conducted the experiment for 14 days. Every day the data-generation phase was executed for around 4 hours followed by the learning phase. The number of interaction steps the robot could perform during 4 hours data-generation phase depended on the internet speed[3] as we used the wireless media for a communication between Pepper and the computer system running MDARQN. For each interaction step, the robot provides eight most recent depth and grayscale frames i.e., $m = 8$. The replay buffer stored up to 3750 most recent interaction experiences. During learning phase, a mini buffer of size 2000 samples was randomly sampled from the replay buffer $\mathcal{M}$. This mini buffer was then used for mini-batch training of the Q-network using RMSProp algorithm. This mini-batch training was repeated 10 times during the learning phase and the size of a mini-batch was 25 samples. As suggested in [8], the initial LSTM hidden and memory state were zeroed for each new mini-batch. The target network parameters were updated every day after training and the learning rate was kept constant at 0.00025. The exploration parameter $\epsilon$ was annealed linearly from 1 to 0.1 over the 28000 interaction steps, however, during 14 days of experiment the robot could perform only 13938 interaction steps due to variations in internet speed at different locations.

[3]With upstream speed of 37 Mbps and downstream speed of 23 Mbps, the robot could execute 2010 interaction steps i.e., $T = 2010$.

| Trained Model | MDARQN(Aug) | MDARQN | MDQN |
|---|---|---|---|
| Hand-shake ratio | 0.74 | - | 0.48 |
| Accuracy (%) | 95.2 | 91.3 | 95.3 |
| True positive rate (%) | 90.5 | 82.7 | 90.7 |
| False positive rate (%) | 3.15 | 5.77 | 3.09 |

TABLE I: Performance measures of trained Q-networks.

### H. Evaluation Procedure

In order to evaluate the MDARQN decisions and the impact of attention model on the human-robot interaction, we carried out two kinds of evaluations:

*1) Evaluating MDARQN decisions on a test dataset:* Since for each given scenario there can be more than one feasible action, therefore, to evaluate either agent decision is right or wrong, we use the following evaluation method. The MDARQN decisions on a test dataset, not seen by the MDARQN during training, were evaluated by three volunteers. The test dataset has 4480 grayscale and depth frames. Each volunteer observes the sequence of eight grayscale frames depicting the scenario followed by the MDARQN decision. The volunteer then decides if the decision is right or wrong. If the decision is marked wrong by the majority of volunteer then the volunteers were asked to pick up the most suitable action for the depicted scenario.

*2) Evaluating the impact of attention mechanism:* We placed the robot in public but this time, the robot interacted with people under our trained Q-networks' policy. The performance of the MDARQN was compared with MDQN through a ratio of number successful handshakes over total number of handshake attempts. The results of the evaluations are presented in the results section.

### I. Source code and data availability

In order to facilitate the implementation of the proposed MDARQN, we release the source code of our complete project together with the depth dataset collected during 14 days of experiment[4]. Although the dataset used for training comprised of both grayscle and depth images but due to privacy concerns, only the depth dataset is made publicly available.

## VI. RESULTS

This section presents the results of the proposed neural Q-networks. Table 1 compares the performance of three models i.e., MDARQN(Aug), MDARQN and MDQN. The MDARQN(Aug) was trained on an augmented training dataset while MDARQN and MDQN were trained on an unaugmented training dataset. The description of nomenclature used in table 1 is as follow. The handshake ratio, as discussed earlier, measures how often the robot augmented with a certain Q-network can attract people for handshaking in an uncontrolled public environment. Accuracy is a measure of how often the Q-network's predictions were correct. True positive rate corresponds to the percentage of predicting

[4]https://sites.google.com/a/irl.sys.es.osaka-u.ac.jp/member/home/ahmed-qureshi/deephri

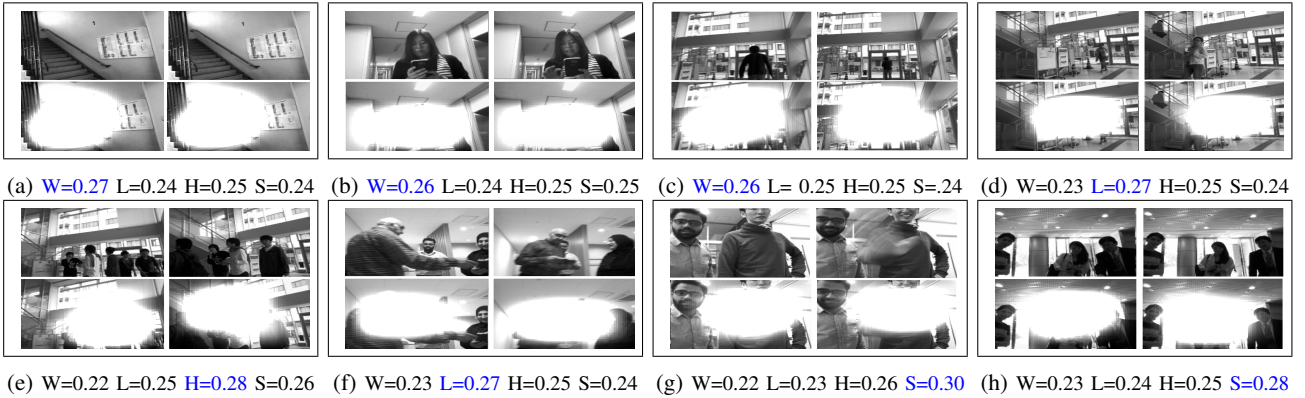

(a) W=0.27 L=0.24 H=0.25 S=0.24    (b) W=0.26 L=0.24 H=0.25 S=0.25    (c) W=0.26 L=0.25 H=0.25 S=.24    (d) W=0.23 L=0.27 H=0.25 S=0.24

(e) W=0.22 L=0.25 H=0.28 S=0.26    (f) W=0.23 L=0.27 H=0.25 S=0.24    (g) W=0.22 L=0.23 H=0.26 S=0.30    (h) W=0.23 L=0.24 H=0.25 S=0.28

Fig. 3: Successful cases of agents decision.

positive targets as positive. False positive rate measures how often the negative examples were classified as positive. The first row of table 1 indicates that the handshake ratios for a robot augmented with MDAQRN(Aug) and MDQN are 0.74 and 0.48, respectively. Furthermore, it can be seen in the last three rows of table 1 that MDARQN(Aug) and MDQN demonstrate similar performance while MDARQN has relatively inferior performance on the test dataset. In addition, we have also noticed that the performance of individual streams of MDQN and MDARQN(Aug) were also similar with a true positive rate of around 70%. This indicates that in order to provide similar performance to the neural network without attention, the attention driven neural networks require more training data.

From now onward all results presented correspond to our proposed MDARQN(Aug). Figures 3 and 4 show the successful and unsuccessful cases of MDARQN(Aug) decisions, respectively, in the depicted scenarios. The actions: wait, look towards human, wave hand and handshake are abbreviated as W, L, H, and S respectively in these figures. In figure 3, in each sub-figure, the top two frames (starting from left) show the first and the last frame out of eight most recent frames for any situation while the bottom two images indicates the region of attention on these frames. An action with maximum Q-value is highlighted in blue to indicate the agent's decision for the particular scenario. The discussion on the MDARQN correct decisions is presented in discussion section. In figure 4, the action highlighted in red is the agent's decision while the action highlighted in green is the decision considered right by the evaluators.

## VII. DISCUSSION

This section provides a brief discussion on the intention prediction ability of MDARQN, impact of attention steering and reward function definition on HRSI.

### A. Intentions depicting factors

As discussed earlier, human intention prediction is crucial for HRSI and human intentions can be predicted from various intention depicting factors. Results in figure 3 indicate that our proposed model has learned to infer intention from those factors. In figure 3(b), an activity is in progress i.e., a person is taking a picture and the agent decides to wait. This action of MDARQN is also in accordance with human social norms as we humans usually do not intervene when someone is taking a picture. Figure 3(c) and 3(d), highlights the ability of our model to interpret human walking trajectory as in the former a person is walking away and agent waits while in the latter, the person is walking towards the robot and the agent decides to look towards the person. Furthermore, our model has also learned to determine the level of human engagement with the robot during an interaction. The scenarios in figures 3(e)-3(h) are arranged in increasing order of human involvement with the robot during the interaction. The scene in figure 3(e) indicates least humans involvement because people are at distance and are not looking towards the robot, the agent takes *wave hand* action to gain people's attention. The scene in figure 3(f) shows relatively higher people involvement so agent chooses *look towards human* action which is a softer way of gaining human attention as compared to wave hand. In the scenarios in figures 3(g)-3(h), people are fully engaged so agent decided to handshake. This level of human engagement with a robot is indicated by person's body orientation and distance from the robot. Hence, the results indicate that the MDARQN has learned to predict human intentions from intention indicating factors.

### B. Impact of attention steering on human-robot interaction

The higher handshake ratio of MDARQN(Aug) as compared to MDQN indicates that people show more willingness to interact with a robot that exhibits it's attention and is responsive to human stimuli compared to a robot that is not. This result is in accordance with the findings of [7] and hence, attentioning is important for a successful HRSI. Despite higher handshake ratio of MDARQN(Aug) with respect to MDQN, the handshake ratio for both models is actually low. One of the reasons for this low ratio is robot's repeated attempts to perform a handshake with a person who is fully engaged (as can also be seen in the accompanying video) but we, humans, avoid multiple handshakes. Therefore, in our future plan, we hope to add memory to the model so that the robot can determine with whom it has already interacted.

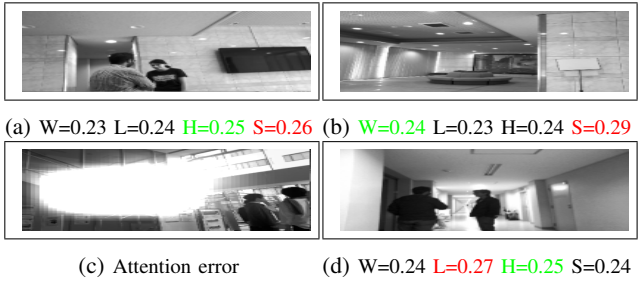

(a) W=0.23 L=0.24 H=0.25 S=0.26   (b) W=0.24 L=0.23 H=0.24 S=0.29

(c) Attention error   (d) W=0.24 L=0.27 H=0.25 S=0.24

Fig. 4: Unsuccessful cases of agents decision.

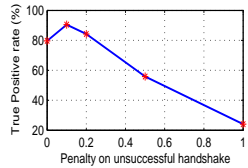

Fig. 5: Effect of reward function on the robot's behavior.

In addition to willingness, the attention is also important to determine with whom the robot is intending to interact out of many other people in the scene. As in figures 3(g) and 3(h), the attention network highlights the person on right side and the person at the center, respectively in order to perform handshake with them. It should be noted that this precise attentioning can not be possible with the attention steering method for non-greedy actions.

### C. Reward function and robot's behavior

Reward function definition determines the robot behavior; the results presented so far were based on the reward function discussed earlier. In this section, we evaluate the effect of different reward functions on the robot's behavior. High penalty on unsuccessful handshake inculcates rude behavior into the robot as robot become reluctant to handshake while low penalty e.g., 0 inculcates amiable behavior as robot repeatedly attempts to do a handshake. In order to test which robot behavior is acceptable, we trained five models and the penalties on unsuccessful handshake for these five models were 0, 0.1, 0.2, 0.5 and 1 while rest of the reward function definitions were kept same as discussed earlier. The performance of these models was evaluated on test dataset following the agent decision evaluation procedure (discussed earlier). The graph in figure 5 shows that model with penalty of 0.1 on unsuccessful handshake generates more socially acceptable decisions as compared to other models.

## VIII. Conclusion

In order to ensure successful HRSI, it is essential for a robot to interpret complex human behavior and respond to these behaviors in a perceivable way. We propose a Multimodal Deep Attention Q-Network (MDARQN) which was trained through a 14 days of hit and trial method based robot interaction with people in real unconstrained public environments. The results of training indicate that our proposed MDARQN enabled the robot to respond to complex human behavior by first interpreting them and then executing a responsive action with attention indication. The results also show i) that the robot has learned to infer intention from intention depicting factors such as human body language, walking trajectory or any ongoing activity; ii) that the attention indication adds perceivability to robot actions and thus people show more willingness for interaction with a robot; iii) the diverse interaction scenarios which were definitely hard to envision and yet the MDARQN learned to choose appropriate decisions in these diverse scenarios.

In our future plan, we plan to i) explore the impact of different fusion strategies on multimodal learning; ii) augment the proposed network with differentiable working memories in order to realize long-term HRI.